\documentclass{article}

\usepackage[nonatbib,preprint]{neurips_2019}

\usepackage[utf8]{inputenc} 
\usepackage[T1]{fontenc}    
\usepackage{hyperref}       
\usepackage{url}            
\usepackage{booktabs}       
\usepackage{amsfonts}       
\usepackage{nicefrac}       
\usepackage{microtype}      
\usepackage{graphicx}

\usepackage{times}
\usepackage{soul}
\usepackage{graphicx}
\usepackage{amsmath}
\usepackage{color}
\usepackage{indentfirst}
\usepackage{multirow}

\title{GazeCorrection:Self-Guided Eye Manipulation in the wild using Self-Supervised Generative Adversarial Networks}

\author{%
  Jichao Zhang \thanks{Equal Contribution} \\
  Shandong University\\
  \And
  Meng Sun \footnotemark[1] \\
  Shandong University \\
  \And
  Jingjing Chen\footnotemark[1] \\
  Shandong University \\
  \And
  Hao Tang \\
  University of Trento \\
  \And
  Yan Yan \\
  Texas State University \\
  \And
  Xueying Qin \\
  Shandong University \\
  \And
  Nicu Sebe \\
  University of Trento \\
}

\begin{document}

\maketitle

\begin{abstract}
Gaze correction aims to redirect the person's gaze into the camera by manipulating the eye region, and
it can be considered as a specific image resynthesis problem. Gaze correction has a wide
range of applications in real life, such as taking a picture with staring at the camera.
In this paper, we propose a novel method that is based on the inpainting model to learn from
the face image to fill in the missing eye regions with new contents representing corrected eye gaze.
Moreover, our model does not require the training dataset labeled with
the specific head pose and eye angle information, thus, the training data is easy to collect.
To retain the identity information of the eye region in the original input,
we propose a self-guided pretrained model to learn the angle-invariance feature.
Experiments show our model achieves very compelling gaze-corrected results in the wild dataset which is collected
from the website and will be introduced in details. Code is available at \url{https://github.com/zhangqianhui/GazeCorrection}.

\end{abstract}

\section{Introduction}

Gaze correction, a specific task of gaze redirection which aims to
 adjust the eye gaze to any directions, just changes the gaze into a single new direction.
In this paper, we focus on the task of gaze correction which is to manipulate the eye gaze to stare at the camera.

In some important life scenarios, these are a need to alter the appearance of eyes in a way that digitally
manipulates the person's gaze into the camera. In same cases, it rarely happens for everyone staring at the camera while taking a group photo. Another scenario is in desktop videoconferencing systems where the eye contact is extremely
 important and the gaze also can express attributes such as attentiveness, confidence, and requirement.
 Unfortunately, eye contact and gaze awareness are lost in most videoconferencing systems. Because videoconferencing
 participants look at their monitors and not directly into the camera. In general, eye correction has practical importance in these application scenarios.

The early research for gaze correction depends on special hardware, for example stereocameras~\cite{criminisi2003gaze,yang2002eye},
kinect sensor~\cite{kuster2012gaze} or transparent mirrors~\cite{kollarits199634,okada1994multiparty}.
Recently, some methods based on machine learning have attained very high-quality synthesized images with corrected gaze, such as DeepWarp~\cite{ganin2016deepwarp}.
DeepWarp uses a deep architecture to directly predict an image-warping flow field with the coarse-to-fine learning process.
However, it requires the paired face dataset labeled with accurate, specific information for head pose and eye angles while
the collection of training image depends on particular hardware. Moreover, DeepWarp would suffer from the failure in the wild images
with large variations in head pose. Another major type is based on the 3D model,
such as GazeDirector~\cite{wood2018gazedirector}. The main idea of GazeDirector is to model the eye region in 3D instead of
trying to predict a flow field directly from an input image. However, it is not easy to model realistically in details.

Compared with the previous methods, our model is simple and novel.
We leverage the image inpainting model with a fully convolutional network as the basis of our model to learn
to fill in the missing eye regions with new contents representing corrected eye gaze.
In the process of training and testing, our inpainting model does not require the data with the specific eye angle
and head pose information. Thus, the training data is easy to collect and use.
Similar to the previous inpainting model~\cite{iizuka2017globally},
we use the global and local discriminator architecture $D$ for performing the adversarial loss to improve the
quality of the inpainted region. Moreover, to preserve the identity information in eye region of the original input,
we propose a Self-Guided Pretrained Model to extract the angle-invariance
features which would be as the input of the network to guide the learning process of the generator and discriminator.
Finally, to improve the quality of inpainted result and stabilize the adversarial learning, we add a self-supervised module for discriminator and generator.
We compare our approach with existing baseline methods for correcting the gaze in the wild image.
Qualitative and quantitative evaluations demonstrate that our model can achieve more compelling results.

Overall, our main contributions include:(1) A simple and novel image inpainting model for gaze correction is proposed. Qualitative and quantitative assessment demonstrate
the effectiveness and superiority of the proposed model. (2) A Self-Guided Pretrained Model to extract the angle-invariance features is proposed and the features would be as an input of the inpainting model
to preserve identity information of the original eye region. (3) A novel self-supervised learning module to improve the quality of inpainted result and stabilize the adversarial training.

\section{Related Work}

{\bfseries Generative Adversarial Networks:}
Generative Adversarial Networks~\cite{goodfellow2014generative} is a powerful implicit generative
model to produce a model distribution that mimics a given target distribution,
and it has been applied to many fields, such as low-level image processing
tasks~(image in-painting~\cite{pathak2016context,IizukaSIGGRAPH2017}, image super-resolution~\cite{Ledig_2017_CVPR,DRIT,wang2018esrgan}),
high-level semantic or style transfer~\cite{pumarola2018ganimation,Zhu_2017_ICCV,tang2019multichannel,Zhang:2018:SGM:3240508.3240594},
video prediction and generation~\cite{wang2018pix2pixHD,chan2018everybody} and even
classical computer vision problem~(Object detection~\cite{Wang_2017_CVPR}).

{\bfseries Image Inpainting:}
Image inpainting, an important task in computer vision and graphics, aims to
fill the missing pixels of an image with plausibly synthesized contents. Most of the previous methods for image inpainting
can be mainly divided into two classes. One is based on traditional diffusion or patch methods with low-level features. For example,
PatchMatch~\cite{barnes2009patchmatch}, a fast nearest neighbor field algorithm, which allows for real-time, high-level image inpainting has been proposed.
In summary, based on low-level features, these traditional methods are not effective to fill in holes on complicated semantic
structures and unable to generate novel objects not found in the source image. The other is the learning-based method.
Recently, CNN-based and GAN-based methods have shown promising performance on image inpainting~\cite{pathakCVPR16context,iizuka2017globally}.
However, we argue that these methods would suffer the limitations for face inpainting, where the inpainting
result does not preserve the identity information of the original input. Similar to the previous
paper~\cite{dolhansky2018eye} which depends on the reference image, we propose a self-guided method to learn
the angle-invariance features which would be the input of the inpainting network to guide the inpainting
process for preserving the identity information.

{\bfseries Eye Gaze Manipulation:}
The previous methods for gaze manipulation can be divided into three classes: 1) hardware-driven,
2) rendering and synthesis, 3) learning-based.

\begin{figure}[htp]
\vspace{-0.3cm}
\begin{center}
\includegraphics[width=1.0\linewidth]{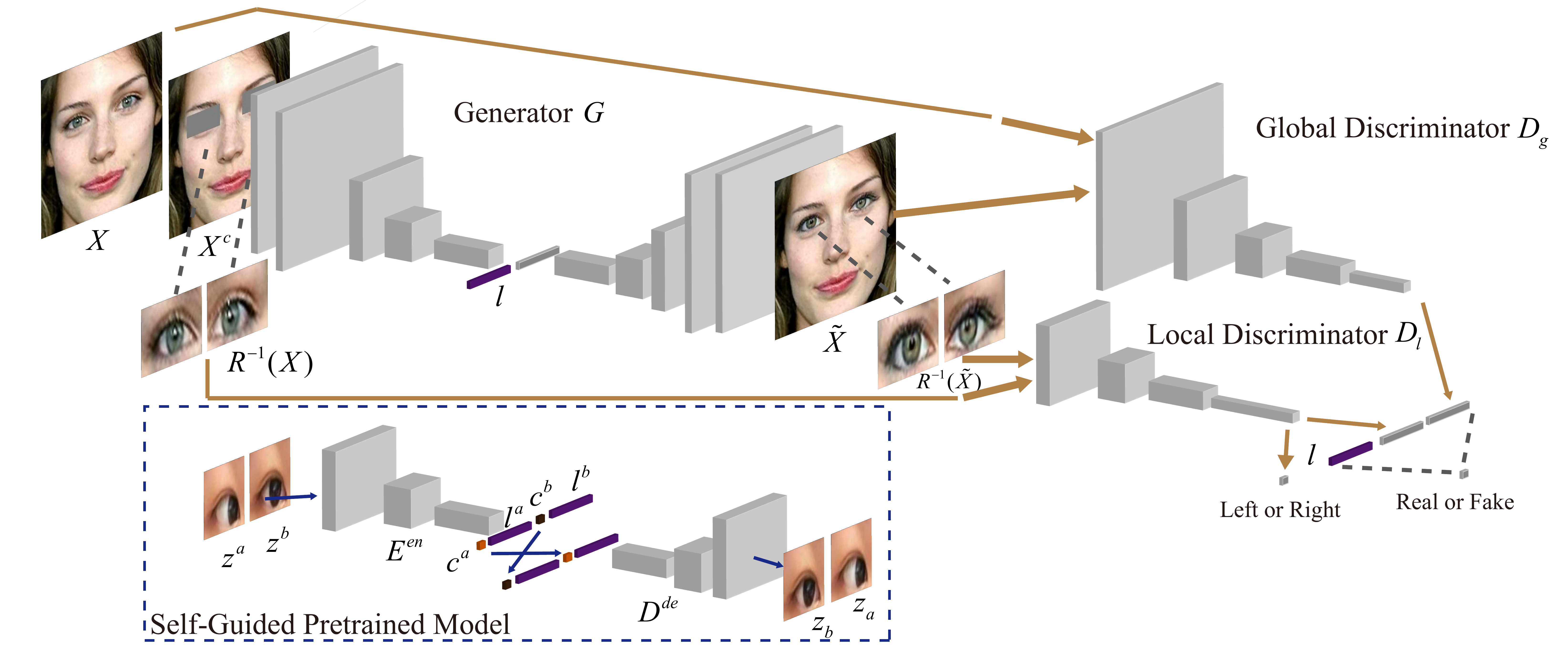}
\end{center}
\vspace{-0.4cm}
\caption{Overview of our network architecture.}
\label{fig:model}
\vspace{-0.5cm}
\end{figure}

The hardware support is indispensable in early research. Kollarits et al.~\cite{kollarits199634} tried to make use of half-silvered mirrors to allow
the camera to be placed on the optical path of the display. Yang et al.~\cite{yang2004eye} aimed to address the eye contract problem with a novel
view synthesis, and they use a pair of calibrated stereo cameras and a face model to track the head pose in 3D. Generally, these hardware-based methods are costly.

Some researchers try to render the eye region based on 3D fitting model, which replaces the original eye with
synthesizing new eyeballs. Banf et al.~\cite{banf2009example} used an example-based approach for deforming the eyelids and slide the iris across
the model surface with texture-coordinate interpolation. To fix the limitation caused by the use of a mesh where the face and eyes are joined, GazeDirector~\cite{wood2018gazedirector}
regards the face and eyeballs as separate parts, synthesizes more high-quality images especially for large redirection angles. These methods for the common gaze
manipulation can be applied for eye correction, but they suffer from the problem where rendering eyes realistically is a challenge.

The core idea for most of the method based on machine learning is to use large training set labeled with specific eye angle and head pose information to
learn the warping field. The warping field is used to relocate eye pixels in the original image, thus realizing gaze redirection. Ganin et al.~\cite{ganin2016deepwarp}
proposed a novel deep model based on the convolutional network as a predictor, which helps them to achieve a high-quality result.
We present a novel method for gaze correction. With an encode-decode architecture as a generator and adversarial process, we can learn from the face image to
fill in the missing regions with new contents representing corrected eye gaze. Compared with those learning-based methods,
our training dataset is unpaired and collected from the website, where the dataset is not labeled with specific eye angle
and head pose information, thus, is easy to collect and use. Furthermore, these models are hard to achieve high-quality results
for gaze correction in the wild images.

\section{Methods}
As shown in Fig.~\ref{fig:model}, our model consists of two parts. One is the generator $G$ is used for inpainting network. We leverage one global discriminator $D_{g}$ with using the entire face as input
and one local discriminator $D_{l}$ with using the local eye region as input to perform the adversarial learning with generator $G$. Additionally, $D_{l}$
 would also perform a self-supervised learning by classifying the right and left eyes.
 The other is a Self-Guided Pretrained model which is based on autoencoder to reconstruct and translate image between different domains to learn the angle-invariance
 features $l$. We will introduce two parts of our model in details.

{\bfseries Generator $G$ Trained with Unpaired Data for the Inpainting:}
We begin with introducing our training data which consists of two domains $\{x_{i}\}^{n}_{1} \in X$ and $\{y_{i}\}^{m}_{1} \in Y$.
Domain $X$: $256 \times 256$ face images with eyes staring at the camera; domain $Y$: $256 \times 256$ face images with eyes staring at somewhere else.
Given a masking function $M$: remove eye region of face data~($M^{-}$: crop out eye region)
and denote the eye-masked face as $X^{c}$ and $Y^{c}$, thus, $X^{c}=M(X)$ and $Y^{c}=M(Y)$. When $n \to \infty$ with the same masking
function $R$, the distribution $P_{X^{c}}$ for $X^{c}$ and $P_{Y^{c}}$ for $Y^{c}$ should be identical, i.e.
$P_{X^{c}}=P_{Y^{c}}$. Given a function $G$ to map $X^{c}$ to $X$, we will have $X=G(X^{c})$, and $G(Y^{c}) \subseteq X$.
This is the theoretical basis of our method.
Therefore, we can use the data from domain $X$ as training data, the data from domain $Y$ as test data. With this mapping function
$G$ trained on domain $X$, we can correct the gaze direction of the test data in the domain $Y$
to stare at the camera. Note that, our training data is unpaired where every instance of domain $X$ does not have the corresponding groundtruth
in the domain $Y$. While the theoretical analysis is still approximately reasonable when the training data is very ample in $X$.

Next, we introduce the learning of generator $G$. As shown in Fig.~\ref{fig:model}, $G$ is designed to be an autoencoder to fill in the missing
regions with the contents representing corrected eye gaze.
Different from the architecture of~\cite{iizuka2017globally}, we use the fully-connected networks instead of
dilated convolution in the bottleneck. For the objective function of generator $G$, a reconstruction loss $L_{rec}$ is necessary, where we use
$L1$ distance between the output $\tilde X~(i.e., G(X^{c}))$ and groundtruth $X$. It is defined as:
\begin{equation} \label{recon_loss}
\begin{aligned}
\ell_{rec} = \mathbb{E}_{X}[\Vert \tilde X - X \Vert_{1}].
\end{aligned}
\end{equation}

{\bfseries Self-Supervised Adversarial Training:}
To avoid the inpainted results tending to be blurry with $\ell_{rec}$ only as an objective function, we employ
two discriminators to perform adversarial loss for improving the visual quality of inpainted result. Additionally, we add a self-supervised learning module
for discriminator to stabilize the training of adversarial networks and further improve the quality of inpainted results. As shown in Fig.~\ref{fig:model}, $D_{g}$  and $D_{l}$ are based on ConvNets. $D_{g}$ takes the entire image $X$ and $\tilde X$ as the input to
ensure the generated contents are semantically coherent on the global region, while $D_{l}$ uses the local
patch $M^{-}(X)$ and $M^{-}(\tilde X)$ as inputs for training to achieve more realistic and sharper contents. We concatenate the final fully-connected output of
$D_{g}$ and $D_l$ into one output which is as the input of sigmoid to predict the probability of the image being real. The objective function
for discriminator $D$~(including $D_{g}$ and $D_{l}$) and generator $G$ is defined as:
\begin{eqnarray}
\mathop{min}\limits_{G}\mathop{max}\limits_{D}\ell_{adv} &=& \mathbb{\mathbb{E}}_{X}[logD(X, M^{-}(X))]
+ \mathbb{\mathbb{E}}_{\tilde X}[log(1-D(\tilde X, M^{-}(\tilde X)))].
\label{eq_gan_loss}
\end{eqnarray}

Inspired by the previous paper~\cite{chen2018self}, we propose a simple and novel self-supervised module. In details,
we simple add a classification learning to divide the left eye and right eye for local discriminator. We desire this additional loss on the image
classification task could improve the ability of representation for local discriminator and stabilize its training. As shown in Fig.~\ref{fig:model}, we augment
the discriminator with a left-right-eye classification loss, which results in the final adversarial loss functions:
\begin{eqnarray}
\mathop{min}\limits_{G}\mathop{max}\limits_{D}\ell_{adv} &=& \mathbb{\mathbb{E}}_{X}[logD(X, M^{-}(X))]
+ \mathbb{\mathbb{E}}_{\tilde X}[log(1-D(\tilde X, M^{-}(\tilde X)))] \nonumber \\ &+& \mathbb{\mathbb{E}}_{X}[logQ_{D}(P=p|X)] - \mathbb{\mathbb{E}}_{\tilde X}[logQ_{D}(P=p|\tilde X)],
\label{eq_gan_loss}
\end{eqnarray}
where $p \in P$ is the position of eye where $p=0$ means left eyes, $p=1$ means right eyes. $Q(P|X)$ is the local discriminator's predictive distribution over the position of eyes.

{\bfseries Self-Guided Pretrained Model for disentangled representation:}
Obviously, this inpainting process is hard to preserve the consistency of the identity information~(e.g., iris color, eye shape) between
$X$ and $\tilde X$. Recently, ExemplarGAN~\cite{dolhansky2018eye}, which is a novel
method for the task of closed-to-open inpainting in natural images, tried to use the exemplar information to produce high-quality, personalized
inpainting results. However, the reference image, generally speaking, is not always available in reality. Inspired by their idea, we can pretrain
a model to attain the angle-invariance features which are as inputs of $G$
and $D$ to guide their training process for preserving the identity information of the eye region.
To learn the angle-invariance features, we denote a  small training data as $\{z^{a}_{i}, z^{b}_{i}\}^{n}_{1} \in Z$, where every image pair $z^{a}_{i}$ and $z^{b}_{i}$ is from the same person,
captured with different angle information in the same scene. As shown in Fig.~\ref{fig:model}, our encoder network $E^{en}$ takes $M^{-}(z^{a})$ and $M^{-}(z^{b})$ as inputs respectively,
the latent codes as the output are $c^{a}, l^{a}=E^{en}(M^{-}(z^{a}))$ and $c^{b}, l^{b}=E^{en}(M^{-}(z^{b}))$~($c=[c_{1},c_{2}] \in \mathbb{R}^{2}$, $l \in \mathbb{R}^{128}$). With the decoder network
$D^{de}$, we firstly reconstruct the original input. The reconstruction objective function
$\ell_{autoencoder}$ for $E^{en}$ and $D^{de}$ is defined as:
\begin{eqnarray} \label{recon_loss_ae}
\ell_{autoencoder} &=& \mathbb{E}_{z^{a},z^{b}}[\Vert M^{-}(z^{a}) - D^{de}(c^{a}, l^{a}) \Vert_{1} + \Vert M^{-}(z^{b})  - D^{de}(c^{b}, l^{b}) \Vert_{1}].
\end{eqnarray}
Then, to learn the translation between $M^{-}(z^{a})$ and $M^{-}(z^{b})$, we swap their latent vector $c$ which is similar to the previous paper~\cite{xiao2018elegant}. The objective function
for the translation is defined as:
\begin{eqnarray} \label{trans_loss_ae}
\ell_{translation} &=& \mathbb{E}_{z^{a},z^{b}}[\Vert M^{-}(z^{b}) - D^{de}(c^{b}, l^{a}) \Vert_{1} + \Vert M^{-}(z^{a})  - D^{de}(c^{a}, l^{b}) \Vert_{1}].
\end{eqnarray}
With $\ell_{autoencoder}+\ell_{translation}$ as the overall objective function for $E^{en}$ and $D^{de}$ trained on paired data,
we can disentangle vector $c$ for representing the angle information in the latent space while vector $l$ can be as the angle-invariance features. \emph{Note that we use the existing gaze dataset to pretrain this model for disentangled representation.}

When training the inpainting model with taking $M^{-}(X)$ as an input of $E^{en}$ for attaining the angle-invariance
features to guide the inpainting of $X$, we call this pretrained model as Self-Guided Pretrained Model and refer to this autoencoder with swapping the latent code as SAE.

{\bfseries Overall Objective function for $G$ and $D$:}
As shown in the Fig.~\ref{fig:model}, the angle-invariance features $l$ is as the input of generator $G$ and discriminator $D$. Taking local eye region $M^{-}(X)$ as the input of $E^{en}$, we have $l=E^{en}_{l}(M^{-}(X))$~($E^{en}_{l}$ means outputing $l$). Then, we would have new $\tilde X~(i.e., G(X^{c}, l))$. Moreover, to measure the distance between the inpainted eye region $M^{-}(\tilde X)$
and $M^{-}(X)$ in perceptual space, we try to minimize $L2$ distance of angle-invariance features $l$ encoded from $M^{-}(\tilde X)$ and $M^{-}(X)$.
In summary, the overall objective functions for $G$ and $D$ are shown as:
\begin{eqnarray}
\mathop{min}\limits_{G}\mathop{max}\limits_{D}\ell_{overall} &=& \mathbb{\mathbb{E}}_{X}[logD(X, M^{-}(X),l)] + \mathbb{\mathbb{E}}_{X}[log(1-D(\tilde X, M^{-}(\tilde X), l))] \nonumber \\
&+& \lambda_{s} \mathbb{\mathbb{E}}_{X}[logQ_{D}(P=p|X)] - \lambda_{s} \mathbb{\mathbb{E}}_{\tilde X}[logQ_{D}(P=p|\tilde X)] \nonumber \\
&+& \lambda_{r}  \mathbb{E}_{X}[\Vert \tilde X - X \Vert_{1}]
+ \lambda_{p} \mathbb{E}_{X}[\Vert E^{en}_{l}(M^{-}(\tilde X)) - l) \Vert^{2}_{2}].
\label{overall_loss}
\end{eqnarray}
The more details about the network architecture have been shown in the supplementary material~\ref{more_result}.

\section{Experiments}
In this section, we first introduce the details of our dataset, network training and baseline models. Next,
we demonstrate the validity of Self-Guided Pretrained Model for learning the angle-invariance features. Then,
we compare the proposed model with baselines by qualitative and quantitative assessment.
Finally, the proposed self-supervised learning module would be investigated.
For brevity, we refer to our method as \emph{GazeGAN}.

\subsection{Dataset}

{\bfseries NewGaze Dataset:}
To evaluate the proposed model, we have investigated the benchmark datasets. However, none of them meet our task for eye correction in the wild.
Thus, we collected a new dataset called NewGaze dataset. NewGaze consists of 30000 images. These unpaired data, which is collected from the CelebA-ID and the website, consists of two domains. Domain $X$: 25000 face images with eyes staring at the camera; domain $Y$: 5000 face images with eyes staring at somewhere else. We crop all images~($256 \times 256$) with face detection algorithm and compute the eye mask region by using facial landmarks detection algorithm.
As described above, we use all data of domain $X$ to train our model, while all data in domain $Y$ is just as the test dataset. Note that no matter paired data is not labeled with the specific eye angle or head pose information, thus, is very easy to collect.

{\bfseries Columbia Dataset:}
The Columbia Gaze~\cite{Smith:2013:GLP:2501988.2501994} is a publicly available gaze data as the benchmark in gaze locking an gaze tracking, where it has 5880 images of 56 people over 5 head pose and 21 gaze directions. For each subjects, they are labelled with three information: head poses~($0^{o}, \pm 15^{o}, \pm 30^{o}$), seven horizontal gaze directions~($0^{o}, \pm 5^{o}, \pm 10^{o}, \pm 15^{o}$) and three vertical gaze directions~($0^{o}, \pm 10^{o}$). The collecting details can be found in the their project page\footnote{http://www.cs.columbia.edu/~brian/projects/columbia\_gaze.html}. Similar to the pre-precessing methods for NewGaze dataset, we would crop and resize them to $256 \times 256$ for training and testing. The training set would be used for training our self-guided pretraining model.
\subsection{Training Details}

\begin{figure}[t]
\vspace{-0.3cm}
\begin{center}
\includegraphics[width=1.0\linewidth]{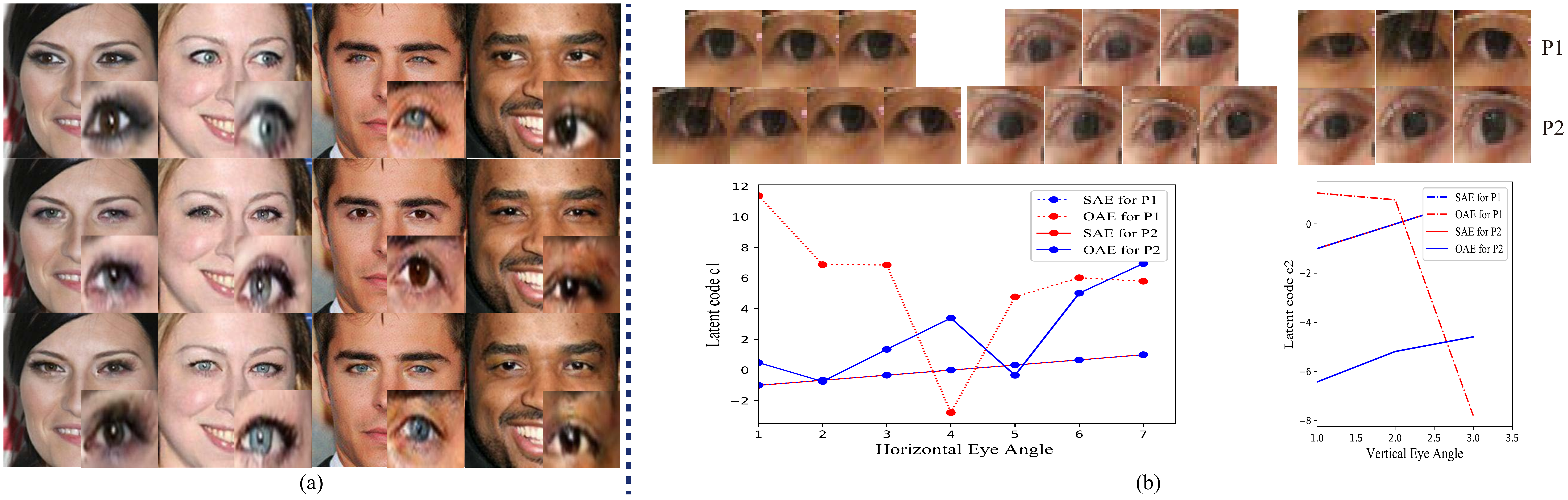}
\end{center}
\vspace{-0.4cm}
\caption{Introduction and validation of disentangled angle representation. (a) Comparisons of the correction result between the GazeGAN(W/O)~(2nd row) and GazeGAN~(3rd row) with the same input~(1st row). The zoomed-in right eyes are shown in the bottom-right for every image. (b) Top: 10 images of eye region with different angles for each person.
Bottom: curves of the latent code $c1$ over eyes with 7 different horizontal direction and curves of the latent code $c2$ over eyes with 3 different
vertical direction. The dotted curves are corresponding to Person 1~(P1) and the solid curves are for Person 2~(P2).}
\label{fig:angle_invariance}
\vspace{-0.3cm}
\end{figure}

The proposed Self-Guided Pretrained model is trained on 1-sized batches with learning rate 0.01.
For the training of main model, $\lambda_{s}$, $\lambda_{r}$ and $\lambda_{p}$ is 1. To stable the training process, we use spectral normalization~\cite{miyato2018spectral}
for every layer in discriminator $D$. The optimizer is Adam with $\beta_{1}=0.5$ and $\beta_{2}=0.999$. The training batch is set to 16.
The learning rate of our inpainting model is 0.0001 for the first 20000 iterations, and it will be linearly decayed to 0 over the remaining iterations.

\subsection{Baseline Models}

{\bfseries Image Inpainting:}
GazeGAN can be regarded as an inpainting model. Thus, we adopt the classical deep inpainting model GLGAN~\cite{iizuka2017globally} as a baseline and train it on the NewGaze dataset.

{\bfseries Image Translation:}
Image translation model, such as, StarGAN~\cite{Choi_2018_CVPR} has achieved high-quality results in facial attribute manipulation.
We train StarGAN on the NewGaze dataset to learn the translation between the domain $X$ and domain $Y$.

{\bfseries Eye Manipulation:}
DeepWarp~\cite{ganin2016deepwarp} has achieved the state-of-the-art gaze manipulation results based on the deep model which uses convolution neural network to predict the flow field for modifying the direction of gaze.
We would compare our model with DeepWarp.

We use the public code of GLGAN\footnote{https://github.com/shinseung428/GlobalLocalImageCompletion\_tf},
and StarGAN\footnote{https://github.com/yunjey/StarGAN}. For DeepWarp without published code,
we implement it by ourselves. DeepWarp requires the paired training image labeled with eye angle and head pose information as inputs. Thus, we train
it using the Columbia gaze dataset~\cite{Smith:2013:GLP:2501988.2501994} and test it in the NewGaze and Columbia dataset.

\subsection{Learning the angle-invariance features}

\begin{figure}[t]
\begin{center}
\includegraphics[width=1.0\linewidth]{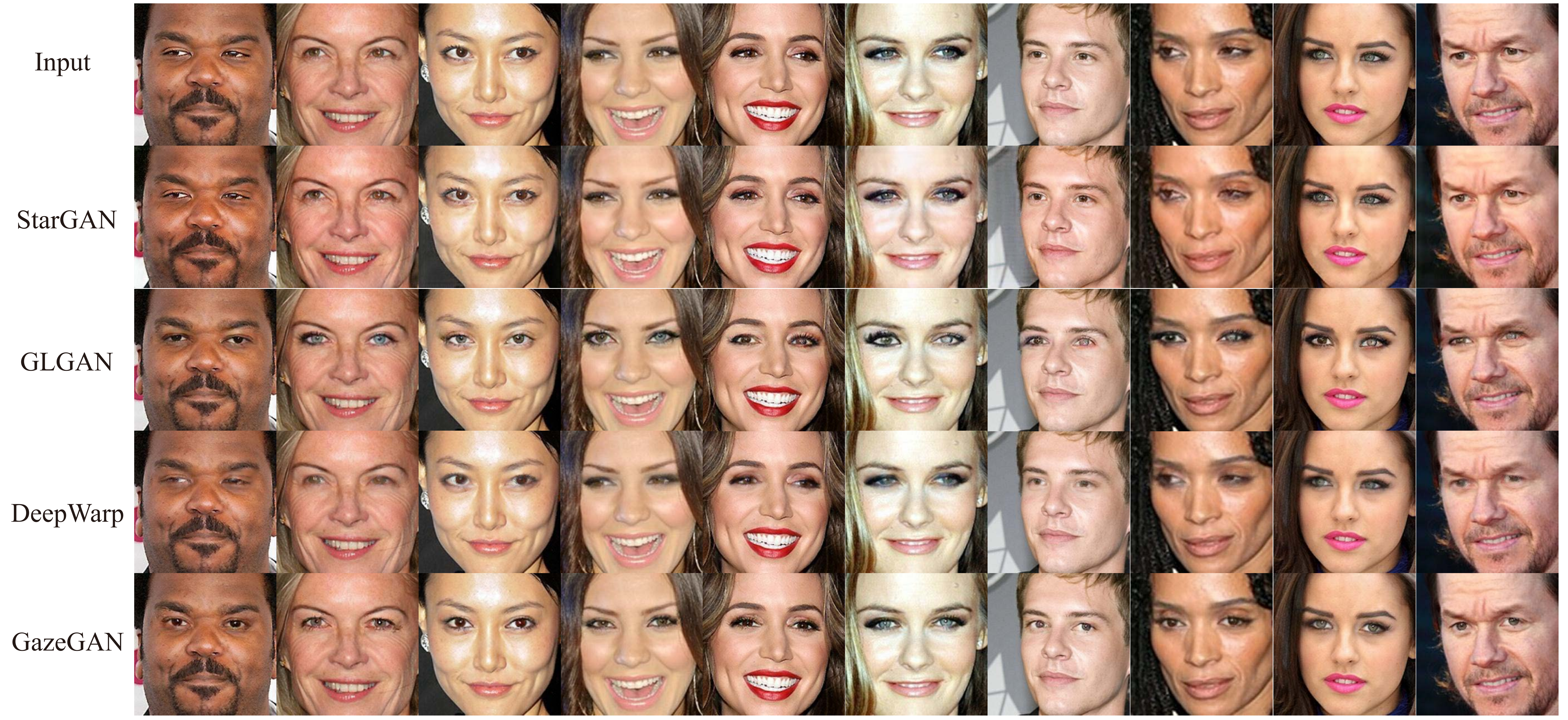}
\end{center}
\vspace{-0.4cm}
\caption{Comparisons with existing works tested in NewGaze dataset. The first row shows the input faces without staring at the camera. The rest rows show the eye-correction results from
different models.}
\label{fig:exp1}
\vspace{-0.3cm}
\end{figure}

We use paired data to train SAE to learn the angle-invariance features $l$, which would be as an input of generator $G$ and
discriminator $D$ to guide the inpainting for retaining the identity information of the eye region in the original input.
We take an opposite approach to validate if $l$ is angle-invariant: when the disentangled features $c$ represents the angle features in the latent space, the remaining features $l$ should be angle-invariant.
Thus, we should validate if the feature $c_{i}$ is a linear correlation with horizontal or vertical eye angle. As shown in left of Fig.~\ref{fig:angle_invariance}(b), we show 7 samples for each
person with different horizontal angles in the top, and give the curve of $c_{1}$ over these 7 samples in the bottom. For original Autoencoder(OAE), we also train it using the paired dataset for comparison.
We can observe that $c_{1}$ encoded from SAE is nearly linear with different eye angles as input while the $c_{1}$ has the similar value for the same angle from different person~(See SAE for P1 and P2). The other curves~(See OAE for P1 and P2) are very irregular means that OAE can not disentangle angle feature in the latent space. As shown in right of Fig.~\ref{fig:angle_invariance}(b), a similar conclusion could be attained.

To further demonstrate the effectiveness of Self-Guided Pretrained model for learning the angle-invariance features,
we train GazeGAN without concatenating the features $l$ as the input of $D$ and $G$, and refer to this variant as GazeGAN(W/O).
As shown in the Fig.~\ref{fig:angle_invariance}(a), the inpainted face for GazeGAN can better preserve the identity information~(iris color and eye shape) of the original face compared with the results of GazeGAN(W/O).

\begin{figure*}[t]
\vspace{-0.4cm}
\begin{center}
\includegraphics[width=1.0\linewidth]{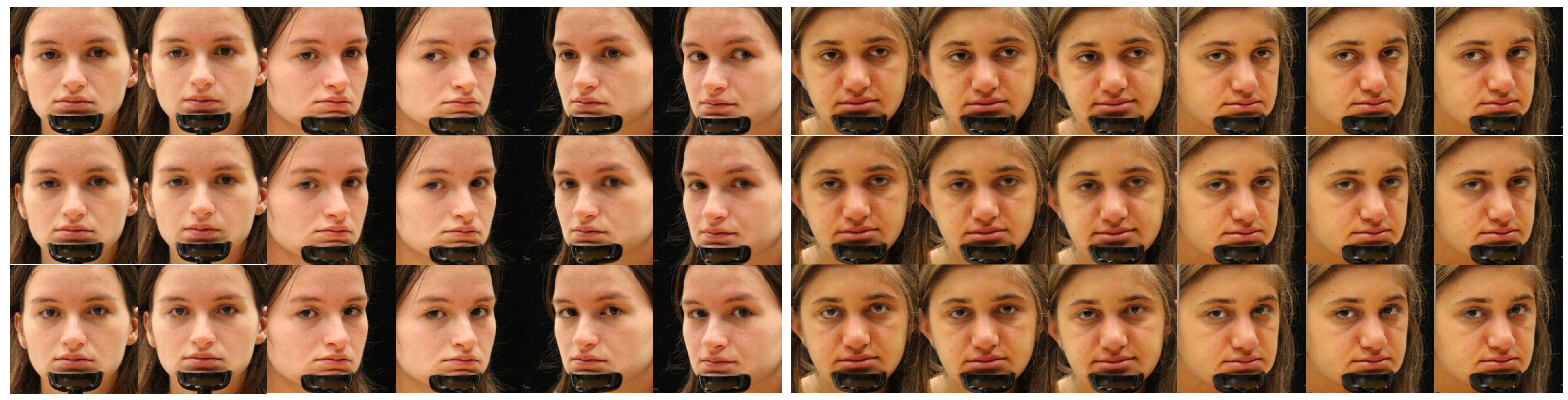}
\end{center}
\vspace{-0.4cm}
\caption{Comparisons of Gaze correction on Columbia test dataset. The rows from top to down are input, results of DeepWarp and GazeGAN.}
\label{fig:exp3}
\vspace{-0.1cm}
\end{figure*}

\begin{figure*}[t]
\vspace{-0.1cm}
\begin{center}
\includegraphics[width=1.0\linewidth]{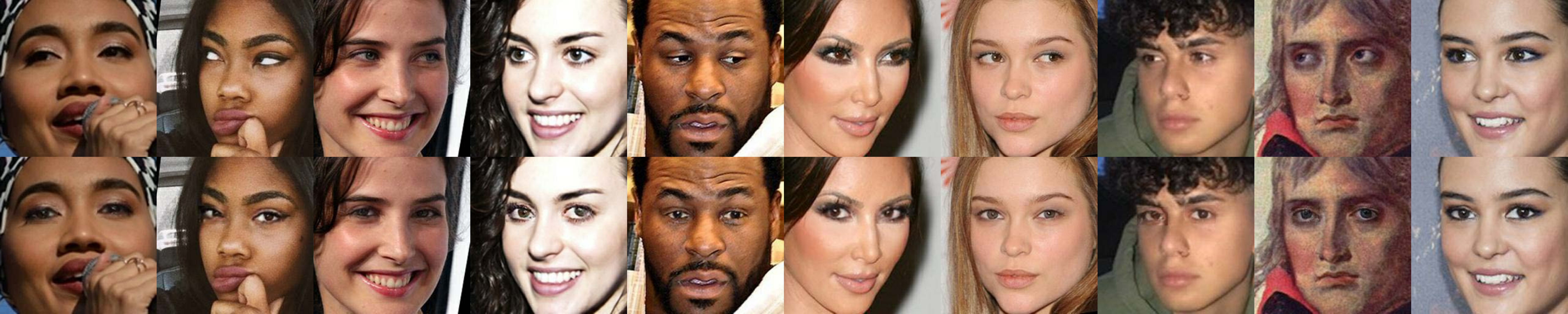}
\end{center}
\vspace{-0.4cm}
\caption{More correction results of GazeGAN in different head pose.}
\label{fig:exp2}
\vspace{-0.6cm}
\end{figure*}

\subsection{Eye correction}

\begin{table}
\vspace{-0.3cm}
\caption{Inception and FID scores on the eye region of gaze-corrected results from different models.}
\begin{center}
\small
\begin{tabular}{cccc}
\hline
Methods & Inception Scores & FID & User Studies~(Average)  \\
\hline
StarGAN & 2.99$\pm$0.08 & {\bfseries 28.34} & 28.90\% \\
 GLGAN & 2.87$\pm$0.07 & 34.33 & 21.87\% \\
DeepWarp & 2.89$\pm$0.12 & 106.53 & 13.13\% \\
GazeGAN & {\bfseries 3.10$\pm$0.12} & 30.21 & {\bfseries35.40\%} \\
\hline
GT   & 3.19$\pm$0.09 & 27.64 & 100\% \\
\end{tabular}
\label{tab:evaluate1}
\vspace{-0.6cm}
\end{center}
\end{table}

\begin{table}
\vspace{-0.3cm}
\caption{Ablation study of self-supervised learning model. The scores are IS and FID metric on 1000 inpainted results generated from models on different training iterations.}
\vspace{-0.3cm}
\begin{center}
\scriptsize
\label{tab:ablation_study}
\begin{tabular}{llllllllllll}
\toprule
Model($10^{4}$) & 2 & 4 & 6 & 8 & 10 & 12 & 14 & 16 & 28 & 20 &\\
\midrule
GazeGAN(W/O) &  2.69       &            2.97 &             2.93 & {\bfseries 3.20} &            2.92 &             3.04 &             2.96 & {\bfseries 3.02} &            3.02  & {\bfseries  3.07} & \multirow{2}{*}{IS} \\
GazeGAN & {\bfseries 3.01} & {\bfseries 3.12} & {\bfseries 2.98} &           3.04 &  {\bfseries 3.12} & {\bfseries 3.10} & {\bfseries 3.10} &           2.98 &  {\bfseries 3.04} & 2.99 &  \\
\midrule
GazeGAN(W/O) &  83.48 & 60.50 &  44.23 & 36.39 &  30.72 &  {\bfseries  28.88} & {\bfseries 29.30} &           34.90 &  34.89 &  34.55 & \multirow{2}{*}{FID} \\
GazeGAN & {\bfseries 53.12} & {\bfseries 37.56} &  {\bfseries 37.51} & {\bfseries 33.36} & {\bfseries 27.86} & 29.15 &  35.21 & {\bfseries 30.00} & {\bfseries 30.19} & {\bfseries 31.74} & \\
\bottomrule
\end{tabular}
\end{center}
\vspace{-0.6cm}
\end{table}

In this section, we provide the comparison experiments with baseline methods on the NewGaze dataset for the task of eye correction, and
use some metrics to evaluate the inpainting results quantitatively. \emph{Note that we do not use any post-processing algorithm for these models, including GazeGAN.}

{\bfseries Qualitative results:}
As shown in the last row of Fig.~\ref{fig:exp1}, GazeGAN can succeed in correcting the eyes into looking at the camera, which validates the effectiveness of the proposed method. In comparison to StarGAN, GazeGAN attains  more
obvious correction results. StarGAN can obtain compelling results in style or texture translation, but it is hard to get the
natural geometric translation. It is explained that StarGAN is based on cycle-consistency loss, which requires the mapping between two domains should be continuous and inverse to each other. According to the invariance of Domain Theorem, the intrinsic dimensions of the two domains should be the same. However, the intrinsic dimensions of domain $Y$ are much higher than domain $X$. Furthermore, compared with GLGAN, GazeGAN demonstrates
its superior advantage in preserving the identity information~(iris color and eye shape) of the original input.
For example, as shown in the 2nd column, the correction result of GazeGAN has a more identical
iris color than GLGAN. We can explain that the angle-invariance features from the proposed Self-Guided Pretrained model contain the identity information from the original input which could guide the inpainting
for preserving the consistency of the color and structure. As shown in the 4th row, the results of DeepWarp are very blurry and do not show an obvious correction for these inputs compared with our model.
Additionally, we test our model and DeepWarp on Columbia test data. The test result has been shown in Fig.~\ref{fig:exp3} and GazeGAN could attain much sharper result than DeepWarp.

As shown in Fig.~\ref{fig:exp2}, we show more high-quality gaze correction results with different head poses.
More results can be found in the supplementary material~\ref{net_arch} or an anonymized project page~\footnote{https://nips2019.wixsite.com/gazegan}.

{\bfseries Quantitative Evaluation Protocol:}
The qualitative evaluation has validated the effectiveness and advantage of our proposed GazeGAN in correcting the eye angle.
To further demonstrate the previous evaluation with quantitative methods, we employ two tactics: Inception scores and FID
to evaluate the generated sample quality in the eye region.

Inception Scores(IS) proposed by~\cite{Salimans:2016:ITT:3157096.3157346} uses the pretrained inception model to compute the
scores on all test results to evaluate the maintenance of meaningful objects and the variety.
Higher IS corresponds to higher quality images. Compared to IS, Fr\'echet Inception distance~(FID) is more consistent
with human evaluation in assessing the realism and variation of input samples. Lower FID means that the input samples are of better quality.

In addition to these two evaluations mentioned above, we conduct a user study in a survey to assess the
results of the eye correction from different models. In details, given an input face image in NewGaze test dataset, we would show the corrected results from different models to 30 respondents who were
asked to select the best one based on perceptual realism and the ability of gaze correction. This study prepares 50 questions for every respondent.

{\bfseries Quantitative results:}
Table~\ref{tab:evaluate1} shows the evaluation results of IS and FID experiments on the corrected images from different models. The proposed
model GazeGAN attains a higher IS and FID scores comparing with GLGAN, e.g., $3.10 \pm 0.12$ for GazeGAN and $2.87 \pm 0.07$ for GLGAN
which demonstrates GazeGAN can produce more photo-realistic inpainting results. We can have a similar conclusion comparing with DeepWarp.  The qualitative evaluation shows StarGAN is hard to learn
the apparent translation, but StarGAN attains a very high score, e.g., 28.34 for FID scores, higher than other methods. We explain that StarGAN can generate high-quality samples and tends to
learn the reconstruction instead of translation by using the Wasserstein GAN with gradient penalty~\cite{gulrajani2017improved} as the objective. The last column of Table~\ref{tab:evaluate1} shows the evaluation results of the user study, and it shows the voting for GazeGAN is $35.40\%$, higher than other models, e.g., $29.60\%$ for StarGAN, $21.87\%$ for GLGAN, $13.13\%$ for DeepWarp.

In general, GazeGAN can achieve very compelling and competitive gaze correction results in the wild image.



\subsection{Ablation Study for Self-Supervised Learning}
We propose a self-supervised adversarial learning module to improve the quality of inpainted result and stabilize the networks for adversarial learning. Our method is very simple and just augment the discriminator and generator with a classification loss for left-and-right eyes. For validating the effectiveness of this self-supervised learning, we conduct the quantitative experiments using IS and FID scores to evaluate the quality of inpainted samples.

As shown in Table.~\ref{tab:ablation_study}, we compare GazeGAN with self-supervised learning and GazeGAN without self-supervised learning module~(GazeGAN(W/O)) over multiple different training iterations. The scores of GazeGAN with self-supervised learning attain higher IS and FID scores in most of cases, which demonstrates the effectiveness of self-supervised learning to improve the quality of inpainted results. In general, our self-supervised learning is very simple, but effective. Moreover, this module is very easy to be augmented into other facial image tasks with adversarial learning.

\section{Conclusion}
In this paper, we have presented a simple and novel model, GazeGAN, for gaze correction in the wild image.
The novelty is we leverage the inpainting model with self-supervised generative adversarial networks
to learn from the face image to fill in the missing eye regions with new contents representing corrected eye gaze.
Moreover, our method is very simple, the proposed GazeGAN does not
require the training data been labeled with specific eye angle and head information,
even the majority of training data without the corresponding groundtruth between different domains.
To preserve the identity information of the original input, Self-Guided Pretrained model proposed could learn the angle-invariance features for guiding
the inpainting process in the training and testing.


\bibliographystyle{plain}
\bibliography{neurips_2019}

\newpage

\appendix
\renewcommand\arraystretch{1.3}
\begin{table*}[htp]
\begin{center}
\Large \textbf{GazeCorrection:Self-Guided Eye Manipulation in the wild using Self-Supervised Generative Adversarial Networks} \\
\vspace{0.2cm}
\normalsize  Supplementary materials are shown below. The network architectures of GazeGAN are shown in Table~\ref{tab:Self-Guided architecture}
,Table~\ref{tab:Generator architecture} and Table~\ref{tab:Discriminator architecture}.  And more qualitative results are shown below.
\end{center}
\normalsize

\section{Network Architecture}

\label{net_arch}
     The Self-Guided network, which is employed to preserve the identity information ,takes as an input the local image rescaled to 128 $\times$ 128 pixels. For the completion network, we use an encoder-decoder architecture,which will add the angle-invariance feature learned by Self-Guided network.And these features are used in discriminator as additional information when determining if the generated image is real or fake. Here are some notations should be noted.  $h$: the height of input images. $w$: the width of input images. C:the number of output channels. K: the size of kernel. S: the size of stride. $P$: the padding method.  IN:instance normalization. FC:Fully-Connected layers.

\footnotesize
\renewcommand\arraystretch{1.5}
\begin{center}

\begin{tabular}{cccccc}

\toprule
Layer & Input Shape & Layer Information & Output Shape\\

\midrule
encoder & $(\frac{h}{2},\frac{w}{2},3)$ & CONV-(C32, K7$\times$7, S1$\times$1,$P_{same}$),IN,ReLU & $(\frac{h}{2},\frac{w}{2},32)$\\
& $(\frac{h}{2},\frac{w}{2},32)$ & CONV-(C64, K4$\times$4, S2$\times$2,$P_{same}$),IN,ReLU & $(\frac{h}{4},\frac{w}{4},64)$\\
& $(\frac{h}{4},\frac{w}{4},64)$ & CONV-(C128, K4$\times$4, S2$\times$2,$P_{same}$),IN,ReLU & $(\frac{h}{8},\frac{w}{8},128)$ \\
& $(\frac{h}{8},\frac{w}{8},128)$ & CONV-(C128, K4$\times$4, S2$\times$2,$P_{same}$),IN,ReLU & $(\frac{h}{16},\frac{w}{16},128)$ \\

\midrule
content & $(\frac{h}{16},\frac{w}{16},128)$ & FC-128& (128,1) \\
rotation & $(\frac{h}{16},\frac{w}{16},128)$ & FC-1 & (1,1) \\
\midrule
decoder & (128,1),(1,1) & Concatenation,FC-($\frac{h}{16} \times \frac{w}{16} \times 128$),reshape & ($\frac{h}{16},\frac{w}{16}$,128)\\
 & ($\frac{h}{16},\frac{w}{16}$,128) &  DECONV-(C128,K4$\times$4,S $\frac{1}{2}$ $\times$ $\frac{1}{2}$ ),IN,ReLU) & $(\frac{h}{8},\frac{w}{8},128)$ \\
&  $(\frac{h}{8},\frac{w}{8},128)$ &  DECONV-(C128,K4$\times$4,S $\frac{1}{2}$ $\times$ $\frac{1}{2}$ ),IN,ReLU) & $(\frac{h}{4},\frac{w}{4},128)$\\
&  $(\frac{h}{4},\frac{w}{4},128)$ &  DECONV-(C64,K4$\times$4,S $\frac{1}{2}$ $\times$ $\frac{1}{2}$ ),IN,ReLU) & $(\frac{h}{2},\frac{w}{2},64)$\\
&  $(\frac{h}{2},\frac{w}{2},64)$  &  DECONV-(C32,K4$\times$4,S $\frac{1}{2}$ $\times$ $\frac{1}{2}$ ),IN,ReLU) & (h,w,32)\\
& (h,w,32) &  CONV-(C3, K7$\times$7, S1$\times$1,$P_{same}$),Tanh & (h,w,3)\\

\bottomrule

\end{tabular}
\end{center}
\vspace{-0.2cm}
\caption{Self-Guided Pretraining architecture.}
\vspace{-0.2cm}
\label{tab:Self-Guided architecture}
\end{table*}

\begin{table*}[htp]
\footnotesize
\begin{center}
\begin{tabular}{cccccc}
\toprule
Part & Input Shape & Layer Information & Output Shape\\
\midrule
encoder & $(h,w,6)$ & CONV-(C16, K7$\times$7, S1$\times$1,$P_{same}$),IN,ReLU &  $(h,w,16)$\\
  & $(h,w,16)$ & CONV-(C32, K4$\times$4, S2$\times$2,$P_{same}$),IN,ReLU & $(\frac{h}{2},\frac{w}{2},32)$\\
        & $(\frac{h}{2},\frac{w}{2},32)$ & CONV-(C64, K4$\times$4, S2$\times$2,$P_{same}$),IN,ReLU & $(\frac{h}{4},\frac{w}{4},64)$ \\
         & $(\frac{h}{4},\frac{w}{4},64)$ & CONV-(C128, K4$\times$4, S2$\times$2,$P_{same}$),IN,ReLU & $(\frac{h}{8},\frac{w}{8},128)$ \\
         & $(\frac{h}{8},\frac{w}{8},128)$ & CONV-(C256, K4$\times$4, S2$\times$2,$P_{same}$),IN,ReLU & $(\frac{h}{16},\frac{w}{16},256)$ \\
         & $(\frac{h}{16},\frac{w}{16},256)$ & CONV-(C256, K4$\times$4, S2$\times$2,$P_{same}$),IN,ReLU & $(\frac{h}{32},\frac{w}{32},256)$ \\

\midrule
bottleneck & $(\frac{h}{32},\frac{w}{32},256)$ & FC-256 &  $(256,1)$ \\
           & $(256,guided_{left}, guided_{right})$ & Concatenation &  $(512,1)$ \\
           & $ (512,1)$ & (FC-256$\times$ $\frac{h}{32}$ $\times$ $\frac{w}{32}$,ReLU) &  $(\frac{h}{32},\frac{w}{32},256)$\\
\midrule
decoder & $(\frac{h}{32},\frac{w}{32},256)$ & DECONV-(C256,K3$\times$3,S $\frac{1}{2}$ $\times$ $\frac{1}{2}$ ),IN,ReLU & $(\frac{h}{16},\frac{w}{16},256)$ \\
        & $(\frac{h}{16},\frac{w}{16},256)$ & DECONV-(C128,K3$\times$3,S $\frac{1}{2}$ $\times$ $\frac{1}{2}$ ),IN,ReLU & $(\frac{h}{8},\frac{w}{8},128)$ \\
        & $(\frac{h}{8},\frac{w}{8},128)$  & DECONV-(C64,K3$\times$3,S $\frac{1}{2}$ $\times$ $\frac{1}{2}$ ),IN,ReLU & $(\frac{h}{4},\frac{w}{4},64)$ \\
        & $(\frac{h}{4},\frac{w}{4},64)$ & DECONV-(C32,K3$\times$3,S $\frac{1}{2}$ $\times$ $\frac{1}{2}$ ),IN,ReLU & $(\frac{h}{2},\frac{w}{2},32)$ \\
        & $(\frac{h}{2},\frac{w}{2},32)$  &DECONV-(C16,K3$\times$3,S $\frac{1}{2}$ $\times$ $\frac{1}{2}$ ),IN,ReLU &
         $(h,w,16)$\\
        & $(h,w,16)$ & CONV-(C3, K7$\times$7, S1$\times$1,$P_{same}$),Tanh & $(h,w,3)$\\
\bottomrule

\end{tabular}
\end{center}
\caption{Generator architecture.}
\label{tab:Generator architecture}
\end{table*}

\renewcommand\arraystretch{1.5}
\begin{table*}[htp]
\scriptsize
\begin{center}
\begin{tabular}{cccccc}
\toprule
Part & Input Shape & Layer Information & output \\
\midrule
$D_{g}$ & $(h,w,3)$ & CONV-(C32, K5$\times$5, S2$\times$2,$P_{same}$), LReLU &  $(\frac{h}{2},\frac{w}{2},32)$ \\
                & $(\frac{h}{2},\frac{w}{2},32)$ & CONV-(C64, K5$\times$5, S2$\times$2,$P_{same}$),LReLU &  $(\frac{h}{4},\frac{w}{4},64)$\\
                & $(\frac{h}{4},\frac{w}{4},64)$ & CONV-(C128, K5$\times$5, S2$\times$2,$P_{same}$),LReLU &  $(\frac{h}{8},\frac{w}{8},128)$\\
              & $(\frac{h}{8},\frac{w}{8},128)$ & CONV-(C256, K5$\times$5, S2$\times$2,$P_{same}$), LReLU &  $(\frac{h}{16},\frac{w}{16},256)$\\
              & $(\frac{h}{16},\frac{w}{16},256)$ &  CONV-(C256, K5$\times$5, S2$\times$2,$P_{same}$), LReLU &   $(\frac{h}{32},\frac{w}{32},256)$\\
              & $(\frac{h}{32},\frac{w}{32},256)$ & CONV-(C256, K5$\times$5, S2$\times$2,$P_{same}$), LReLU & $(\frac{h}{64},\frac{w}{64},256)$ \\
               & $(\frac{h}{64},\frac{w}{64},256)$ & FC-256 & (256,1)\\

\midrule
$D_{l}$ & $(\frac{h}{2},\frac{w}{2},3)$,$(\frac{h}{2},\frac{w}{2},3)$ & Concatenation & $(\frac{h}{2},\frac{w}{2},6)$ \\
& $(\frac{h}{2},\frac{w}{2},6)$ & CONV-(C32, K5$\times$5, S2$\times$2,$P_{same}$), Leaky ReLU &  $(\frac{h}{4},\frac{w}{4},32)$ \\
                & $(\frac{h}{4},\frac{w}{4},32)$ & CONV-(C64, K5$\times$5, S2$\times$2,$P_{same}$),LReLU &  $(\frac{h}{8},\frac{w}{8},64)$\\
                & $(\frac{h}{8},\frac{w}{8},64)$ & CONV-(C128, K5$\times$5, S2$\times$2,$P_{same}$),LReLU &  $(\frac{h}{16},\frac{w}{16},128)$\\
              & $(\frac{h}{16},\frac{w}{16},128)$ & CONV-(C256, K5$\times$5, S2$\times$2,$P_{same}$), LReLU &  $(\frac{h}{32},\frac{w}{32},256)$\\
              & $(\frac{h}{32},\frac{w}{32},256)$ &  CONV-(C256, K5$\times$5, S2$\times$2,$P_{same}$), LReLU &   $(\frac{h}{64},\frac{w}{64},256)$\\
              & $(\frac{h}{64},\frac{w}{64},256)$ & FC-256 & (256,1)\\
              & $(\frac{h}{64},\frac{w}{64},256)$ & FC-2 & (2,1)\\
\midrule
             & $(global,local_{left},local_{right}, guided_{left},guided_{left})$&Concatenation,FC-256,ReLU& (256,1) \\
             & (256,1) & FC-1 & (1,1)\\

\bottomrule
\end{tabular}
\end{center}
\caption{Global discriminator $D_{g}$ and Local discriminator $D_{l}$ architecture.}

\label{tab:Discriminator architecture}
\end{table*}

\begin{figure*}
\section{Additional Qualitative Results} \label{more_result}
\begin{center}
\includegraphics[width=1.0\linewidth]{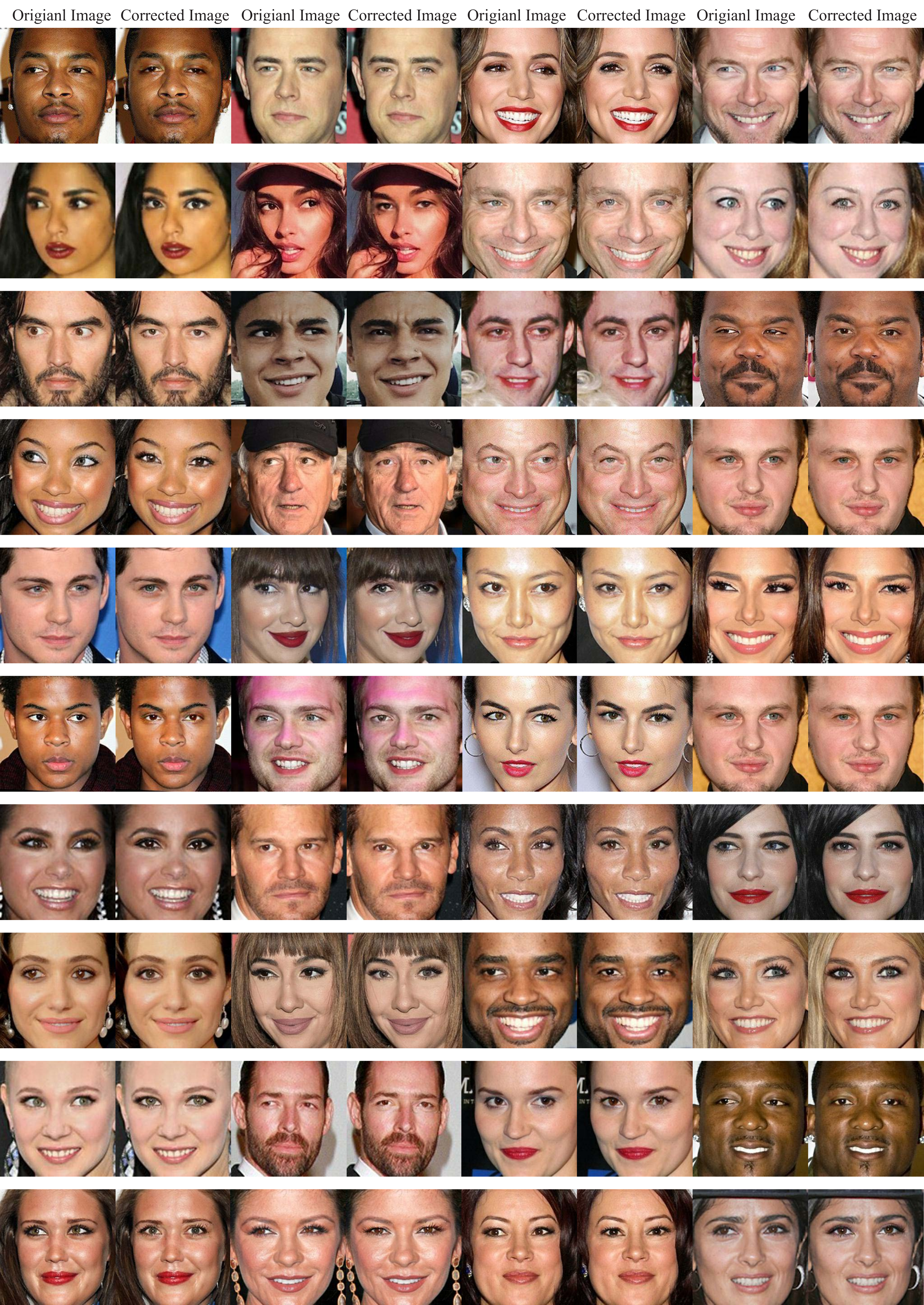}
\end{center}
\vspace{-0.2cm}
\caption{More results of GazeGAN for gaze correction in the wild.}
\label{fig:more_results1}

\end{figure*}

\begin{figure*}
\vspace{-0.2cm}
\begin{center}
\includegraphics[width=1.0\linewidth]{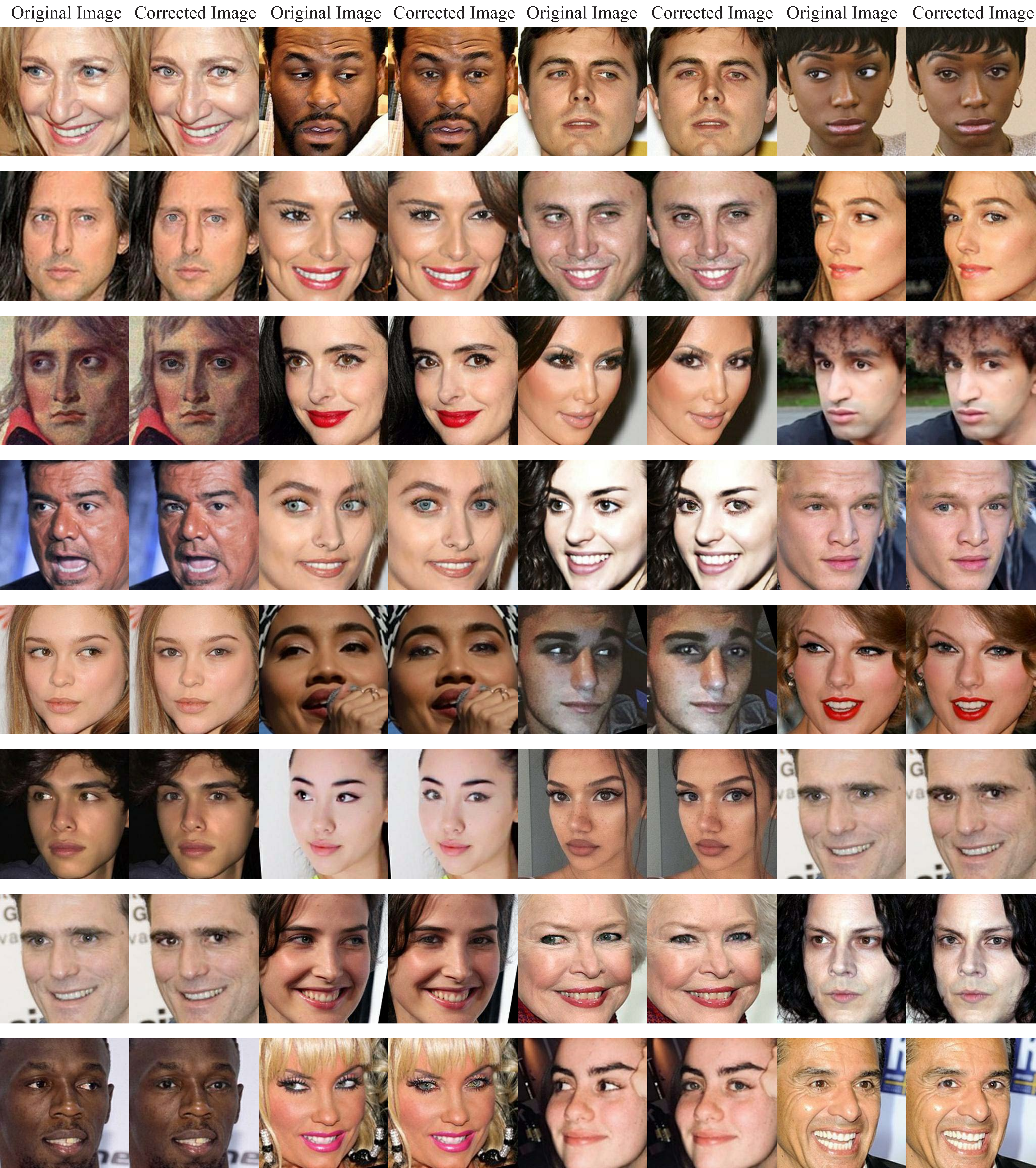}
\end{center}
\vspace{-0.4cm}
\caption{More results of GazeGAN for gaze correction in the wild with large pose.}
\label{fig:more_results2}

\end{figure*}

\end{document}